\documentclass{egpubl}
\usepackage{EGUKCGVC2022}
 
\WsPaper           

\usepackage[T1]{fontenc}
\usepackage{dfadobe}  

\biberVersion
\BibtexOrBiblatex
\usepackage[backend=biber,bibstyle=EG,citestyle=alphabetic,backref=true]{biblatex} 
\electronicVersion
\PrintedOrElectronic
\ifpdf \usepackage[pdftex]{graphicx} \pdfcompresslevel=9
\else \usepackage[dvips]{graphicx} \fi

\usepackage{egweblnk} 

\usepackage{array}
\newcolumntype{C}[1]{>{\centering\arraybackslash}p{#1}}
\usepackage{lipsum}
\usepackage{todonotes}
\usepackage{url}

\newcommand{\etal}{\emph{et al}. }
\usepackage{tikz}
\usepackage{comment}
\usepackage{amsmath,amssymb} 
\usepackage{color}

\title[Depth-aware Neural Style Transfer using Instance Normalization]%
      {Depth-aware Neural Style Transfer using Instance Normalization}

\author[E. Ioannou \& S. Maddock]
{\parbox{\textwidth}{\centering E. Ioannou\orcid{0000-0003-3892-2492} and S. Maddock\orcid{0000-0003-3179-0263}}
        \\
{\parbox{\textwidth}{\centering Department of Computer Science, The University of Sheffield, UK
         }
}
}

\begin{document}

\teaser{
    \centering
    \includegraphics[width=\linewidth]{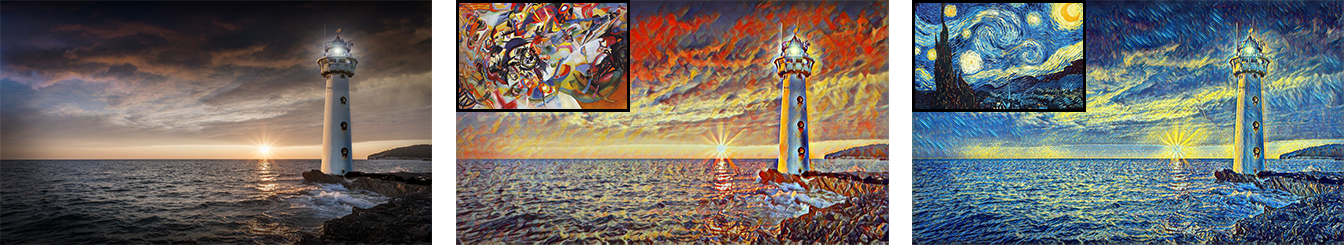}
    \caption{Results using our approach: The colour and texture patterns of the style image are captured while depth information is preserved.}
\label{fig:teaser}
}

\maketitle
\begin{abstract}
Neural Style Transfer (NST) is concerned with the artistic stylization of visual media. It can be described as the process of transferring the style of an artistic image onto an ordinary photograph. Recently, a number of studies have considered the enhancement of the depth-preserving capabilities of the NST algorithms to address the undesired effects that occur when the input content images include numerous objects at various depths. Our approach uses a deep residual convolutional network with instance normalization layers that utilizes an advanced depth prediction network to integrate depth preservation as an additional loss function to content and style. We demonstrate results that are effective in retaining the depth and global structure of content images. Three different evaluation processes show that our system is capable of preserving the structure of the stylized results while exhibiting style-capture capabilities and aesthetic qualities comparable or superior to state-of-the-art methods. Project page: \url{https://ioannoue.github.io/depth-aware-nst-using-in.html}.

\ccsdesc[500]{Computing methodologies~Image processing}
\ccsdesc[300]{Computing methodologies~Image representations}
\ccsdesc[300]{Applied computing~Fine arts}
\ccsdesc[300]{Applied computing~Media arts}

\printccsdesc   
\end{abstract}

\section{Introduction}

Neural Style Transfer (NST) is concerned with the artistic stylization of various forms of data, such as images, videos and 3D models. In the context of 2D image stylization, which is where NST has predominantly been applied, this can be described as the process of transferring the style of one image onto an input ‘content' image.  The technique, which has attracted wide attention in academia and industry, is capable of mapping the style patterns of an artistic image onto an ordinary photograph, synthesizing a novel image that preserves the contents of the photograph while embodying the artistic influences of the particular artwork. 

Many studies have extended the seminal 2D work of Gatys~\etal~\cite{gatys2016image} to other media, such as 3D images, videos, and games \cite{ruder2018artistic,huang2017real,gao2018reconet,deliot_guinier_vanhoey_2020}. Multiple studies have also addressed the computational complexity, speed or aesthetics and the visual quality of the stylized results \cite{johnson2016perceptual,ulyanov2016texture,sanakoyeu2018style,hu2020aesthetic}. Recently, a considerable amount of research has been devoted to enhancing the structure and depth-preserving capabilities of the NST algorithms based on the observation that the stylized images often neglect much of the content information by applying the style patterns evenly throughout the whole image \cite{liu2017depth,cheng2019structure,kitov2019depth}.
To eliminate these undesired effects, which are especially visible when the input content images include objects at various depths, algorithms have been proposed that in addition to content and style loss also encompass a depth reconstruction loss in training \cite{liu2017depth,cheng2019structure}. This is achieved by utilizing state-of-the-art depth estimation approaches \cite{chen2016single}.

We present an approach that is based on the image transformation network introduced by Johnson~\etal~\cite{johnson2016perceptual} and produces stylized results that preserve the global structure and depth of the contents. Our algorithm uses Instance Normalization (IN) layers instead of Batch Normalization (BN), a modification to style transfer approaches proposed by Ulyanov~\etal \cite{ulyanov2016instance} that improves the quality of the results. In addition, we utilize a state-of-the-art depth estimation network \cite{Ranftl2020,Ranftl2021} for the computation of the depth information that serves as an additional loss function to the content and style losses. This method extends the method by Liu~\etal \cite{liu2017depth} which initially introduced the idea of incorporating depth reconstruction loss for the training of the image transformation network as a way of generating stylized results that take into account the depth of the input content images. We show that our approach, by making use of a more accurate depth estimation approach than Liu~\etal, and also by replacing the Batch Normalization layers with Instance Normalization, is capable of producing results that better embody the style contrast across different areas of the image, thus improving upon the quality of the stylization.

The rest of the paper is organized as follows. Section~\ref{sec:RelatedWork} presents the related work. Our method is introduced and analyzed in detail in Section~\ref{sec:Method}. Section~\ref{sec:Results} contains the results of our approach along with a discussion about its effectiveness. Finally, Section~\ref{sec:Conclusions}, supplies conclusions and discussion about future work.


\section{Related work}
\label{sec:RelatedWork}

Empowered by the comparable-to-human capabilities of Convolutional Neural Networks (CNNs) in object recognition, Gatys~\etal suggested a system that reproduces famous paintings on natural images \cite{gatys2016image}. The algorithm takes as input a content image and a style image and initializes a noise image which is subsequently optimized by seeking to minimize an objective function that encompasses definitions of content loss and style loss. The content is represented by the higher-level features of a pre-trained \textit{VGG-19} network \cite{simonyan2015deep} while the style is considered as a set of summary statistics. For the representation of the style, features are extracted from multiple layers and feature correlations are calculated with the use of Gram matrices. Despite the effectiveness and sophistication of this procedure, it requires a notable amount of time to generate a single stylized image. The method of Johnson~\etal~\cite{johnson2016perceptual} avoids this slow optimisation process and proposes an algorithm that utilizes perceptual loss functions to train their networks. A generative model is optimized offline allowing the stylized output to be produced with a single forward pass, which is orders of magnitude faster. Similar algorithms manage to improve upon the speed and overall computational cost by learning feed-forward networks \cite{ulyanov2016texture,ulyanov2017improved}, while further work considers the incorporation of multiple styles per model \cite{dumoulin2016learned,chen2017stylebank} or even an arbitrary style per model \cite{huang2017arbitrary,chen2016fast,gu2018arbitrary}.

Further improvements to earlier approaches were shown by using instance normalization (IN) -- or contrast normalization -- instead of batch normalization \cite{ulyanov2017improved}. Similarly to Johnson~\etal \cite{johnson2016perceptual}, Ulyanov~\etal~\cite{ulyanov2016texture} use a generator that is composed of convolutions, pooling, upsampling and batch normalization. Ulyanov~\etal~\cite{ulyanov2016instance} suggest the same configuration but with contrast normalization layers in order to prevent the stylized results from depending on the contrast of the content image. The results demonstrate better quality and inhibit the undesired distribution of style patterns across the whole image \cite{ulyanov2017improved}. 

Other approaches view the problem from a different perspective, trying to redefine what is considered to be the “style” of an artwork. Such attempts include the algorithm of Sanakoyeu~\etal~\cite{sanakoyeu2018style} which trains the network to focus on the details that are relevant for the style when measuring the similarity in content between the input and the stylized image, resulting in a more generalized procedure that avoids fixed style representations (captured by the features of a pre-trained \textit{VGG} network). Another method presented by Hu~\etal~\cite{hu2020aesthetic} aims to control the aesthetic of the stylized result by separately manipulating the colour and texture features. Their system inputs two different images that reference colour and texture instead of only one style image. These approaches seem to exceed the limitation of using a single style image and improve upon the aesthetic qualities of the stylized results, offering a viewpoint that, unlike previous studies, focuses on the style image's characteristics and overall aesthetics.

Most of the aforementioned studies neglect to consider depth preservation and coherence of details, yet, depth information is considered to be of great importance when evaluating the visual quality of the NST methods’ results \cite{jing2019neural}. To address this limitation, Liu~\etal~\cite{liu2017depth} proposed a system that is based on the work of Johnson~\etal~\cite{johnson2016perceptual} but integrates depth estimation. Their method suggests the addition of depth reconstruction loss in training of the transformation network and makes use of a single-image depth perception network \cite{chen2016single}. An extension to this work was implemented by Cheng~\etal~\cite{cheng2019structure}, whose approach focuses on retaining or enhancing the structure of the artistically stylized result. Using a global structure extraction network (represented by the depth map) and a local structure refinement network (represented by the image edges), they provide an adjustable way to control the amount of structure that is preserved when stylizing an image. This results in stylized outputs that do not suffer from style textures being scattered over the whole image and disrupting the content structures, which is aesthetically better, especially when the content image contains a face or multiple objects at various depths. However, as the authors suggest, the method might be unsuitable for users that prefer a more abstract feeling. Another approach by Kitov~\etal~\cite{kitov2019depth}, which is based on the method of Huang~\etal~\cite{huang2017arbitrary}, applies stylization in different regions of the content image with different strength depending on their distance to the camera. Our approach focuses on retaining the global structure of an image, building on the initial method of Liu~\etal~\cite{liu2017depth}. 

Previous approaches rely on depth estimation methods to calculate depth from a 2D image. Predicting depth from a single RGB image is a long-standing problem in computer vision. The earliest data-driven methods that emerged with the rise of Deep Learning mostly make use of neural networks that are trained on ground-truth metric depth. 

The recent work by Ranftl~\etal \cite{Ranftl2020,Ranftl2021}, relies on the idea that effective monocular depth estimation is tightly dependent on the variety and diversity of the training data. The method consists of a supervised model that is trained on five different diverse datasets, taking into account indoor and outdoor scenes including static and dynamic objects in various contexts. As part of our contributions, we compare these state-of-the-art methods on depth estimation and utilze a depth estimation network for the training of our models that incorporates a depth reconstruction loss. 


\section{Method}
\label{sec:Method}

\subsection{Overview}
\label{sec:method_overview}

\begin{figure}
	\centering
    \includegraphics[width=\linewidth]{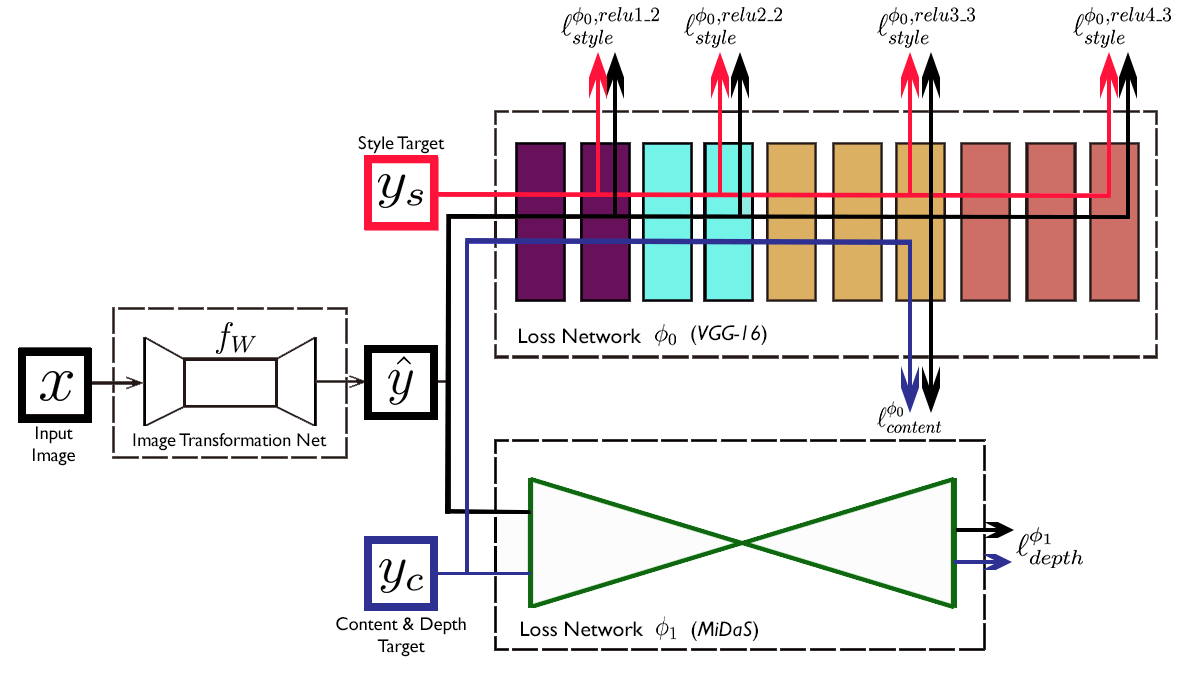}
    \caption{System Overview: The Image Transformation Network which transforms an input image $x$ into an output image $\hat{y}$ is trained. A pre-trained image classification network (\textit{VGG-16}) is used to define the content and style losses. For the definition of depth loss, we use a state-of-the art depth estimation network (\textit{MiDaS}). To generate a style transfer model, for a particular style, the total loss (content loss + style loss + depth loss) is being optimized during training.}
    \label{fig:system_overview}
\end{figure}

\begin{figure}
	\centering
    \includegraphics[width=\linewidth]{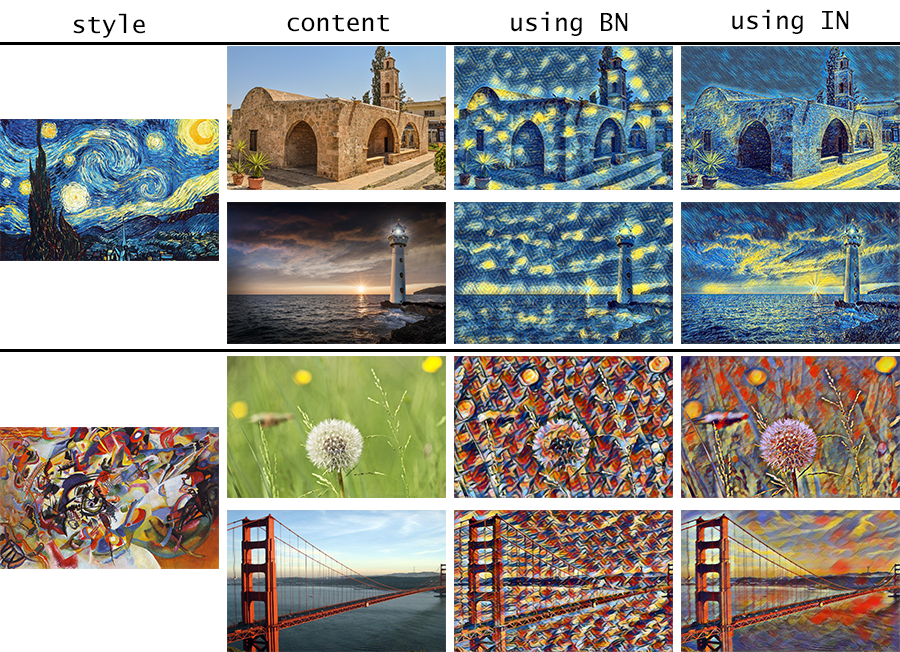}
    \caption{Comparison of the results of the system proposed by Johnson~\etal \cite{johnson2016perceptual} when using Batch Normalization (BN) and Instance Normalization (IN)}
    \label{fig:johnsonetal-in-vs-bn}
\end{figure}

Figure~\ref{fig:system_overview} provides an overview of the overall architecture used in our approach. Similarly to Liu~\etal~\cite{liu2017depth}, our method uses an image transformation network ($f_{W}$) -- a deep residual convolutional network -- that transforms an input image $x$ into an output image $\hat{y}$ via the mapping $\hat{y}=f_{W}(x)$. Unlike Liu~\etal~\cite{liu2017depth}, we use Instance Normalization (IN) layers instead of Batch normalization, so that normalization is applied to single images instead of a whole batch of images. This is based on the observation of Ulyanov~\etal \cite{ulyanov2016instance} that such a modification makes the network agnostic to the contrast of the original images by preventing instance-specific mean and covariance shift. The final configuration of our network thus consists of (i) the model used by Johnson~\etal \cite{johnson2016perceptual}, but with IN layers, and (ii) a depth perception network that is used to capture the depth loss. Figure~\ref{fig:johnsonetal-in-vs-bn} demonstrates the improvements in the aesthetic of the stylized results generated when the network proposed by Johnson~\etal is configured with IN layers instead of Batch Normalization. The original method of Johnson~\etal discards most of the content information, applying style patterns evenly throughout the whole 2D image, whereas replacing Batch Normalization with IN favours structure and content preservation. Therefore, it is sensible to configure our architecture with IN layers since we aim to pay attention to depth and structure information and produce results with better style contrast.

\subsection{Content \& Style Losses}
\label{sec:method_content_style}

In addition to the image transformation network, we also use two loss networks to capture three different losses: a pre-trained image classification network to capture content loss and style loss, and a depth prediction network to capture depth loss. As with Johnson~\etal, we use \textit{VGG-16} \cite{simonyan2015deep}) and its high-level features in order to define the content and style losses. Based on the observation that the deeper layers of a pre-trained convolutional network transform the input image into feature maps that increasingly care about the content of the image rather than any detail about the texture or colour of pixels, the content loss is defined by the squared Euclidean distance between the feature representations of the content image and the transformed image at a particular layer of the network ($relu2\_2$):

\begin{equation} \label{eq:contentloss}
    l^{\phi_{0}}_{content} (\hat{y}, x) = \frac{1}{C_j H_j W_j}  \| \phi^{j}_{0}(\hat{y}) - \phi^{j}_{0}(x) \|^{2}_{2}
\end{equation}
where $\phi_{0}$ is the image classification network and $\phi^{j}_{0}$ represents the activations of the $j^{th}$ layer of $\phi_{0}$ when processing an image with shape $H \times W \times C$ where $H$ denotes the height, $W$ the width and $C$ the number of channels.

For the calculation of the style loss, features are extracted from multiple layers and the feature correlations are given by the Gram matrix $G$ which contains non-localized information about the image:

\begin{equation} \label{eq:gram-matrix}
    G^{\phi_{0}}_{j} (x)_{C, C'} = \frac{1}{C_j H_j W_j}  \sum^{H_{j}}_{h=1} \sum^{W_{j}}_{w=1} \phi^{j}_{0}(x)_{h,w,c}  \phi^{j}_{0}(x)_{h,w,c'}
\end{equation}

The style loss is defined by the squared Frobenius norm between the Gram-based style representations of the transformed image $\hat{y}$ and style image $y$:
\begin{equation} \label{eq:style-loss}
    l^{\phi_{0}, j}_{style} (\hat{y}, y)  = \|  G^{\phi_{0}}_{j} (\hat{y}) - G^{\phi_{0}}_{j} (y) \|^{2}_{F}
\end{equation}

The total style loss is then defined as: 
\begin{equation} \label{eq:style-loss-total}
    l^{\phi_{0}}_{style} (\hat{y}, y)  = \sum_{j \in J} l^{\phi_{0}, j}_{style} (\hat{y}, y)
\end{equation}
where J= \{ $relu1\_1$, $relu2\_2$, $relu3\_3$, $relu4\_3$ \}  is the set of selected layers.

\subsection{Depth Loss}
\label{sec:method_depth}

\begin{figure}[!htb]
	\centering
    \includegraphics[width=\linewidth]{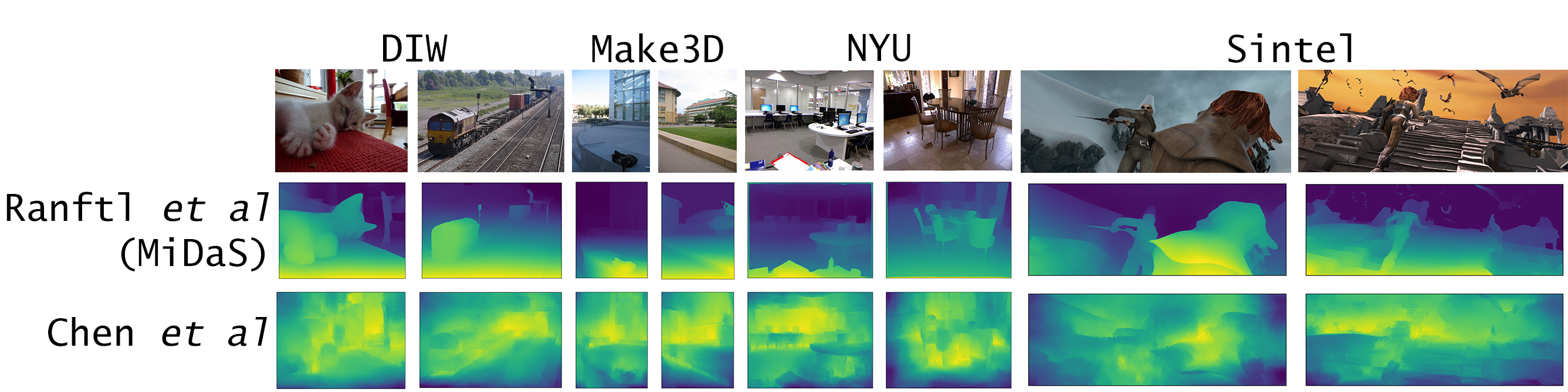}
    \caption{Visual qualitative comparison of the depth estimation methods by Ranftl~\etal (\textit{MiDaS}) \cite{Ranftl2020,Ranftl2021} and Chen~\etal \cite{chen2016single}}
    \label{fig:depth-methods-comparison}
\end{figure}

The superiority of \textit{MiDaS} \cite{Ranftl2020,Ranftl2021} over the pre-existing single-image depth estimation methods guided our choice for the depth network that is utilized in order to compute the depth loss. In their work, they report an overall better performance in comparison with the system proposed by Chen~\etal \cite{chen2016single}. Figure~\ref{fig:depth-methods-comparison} provides a visual comparison between the two methods applied on a variety of images. The images are sampled from a set of diverse datasets, including \textit{DIW} \cite{chen2016single}, \textit{NYU} \cite{silberman2012indoor}, \textit{Make3D} \cite{saxena2008make3d} and \textit{Sintel} \cite{Butler2012sintel}. We use images from the test datasets which were not used during the training of either of the two algorithms.

The chosen depth estimation network ($\phi_{1}$) takes as input an image and directly calculates the depth map. The depth loss is thus defined as the Euclidean distance between the responses of the depth estimation network in regard to the original content image and the transformed image:

\begin{equation} \label{eq:depth-loss-total}
    l^{\phi_{1}}_{depth} (\hat{y}, x)  = \frac{1}{C_j H_j W_j}  \| \phi_{1}(\hat{y}) - \phi_{1}(x) \|^{2}_{2}
\end{equation}

\subsection{Training Details}

The algorithm is trained on the Microsoft COCO dataset \cite{lin2014microsoft} which consists of 80k images. For the training, each image is resized to $256\times256$. We use Adam optimizer \cite{kingma2014adam} with a learning rate of $1 \times 10^{-3}$ and train with a batch size of 4. As discussed previously, the content loss is computed at the $relu2\_2$ layer and the style reconstruction loss at layers $relu1\_2$, $relu2\_2$, $relu3\_2$ and $relu4\_3$ of the \textit{VGG-16} loss network. The depth reconstruction loss is computed at the output layer of the \textit{MiDaS} \cite{Ranftl2020,Ranftl2021} network. We found the optimal weights for the content, style and depth loss to be $1 \times 10^{5}$, $1 \times 10^{10}$ and $1 \times 10^{3}$, respectively. The source code is available from the project's webpage: \url{https://ioannoue.github.io/depth-aware-nst-using-in.html}.


\section{Results and Discussion}
\label{sec:Results}


We compare the results of our method against state-of-the-art approaches \cite{gatys2016image,johnson2016perceptual,liu2017depth} both qualitatively and quantitatively. We perform a side-to-side visual comparison, we present the results of a user study designed to capture the subjectivity when evaluating the aesthetics of the results, and also consider quantitative metrics that assess the capability of the algorithms to preserve the content's depth and enhance the style contrast across the image.

\subsection{Comparison with state-of-the-art methods}

\begin{figure}[!htb]
	\centering
        \includegraphics[width=\linewidth]{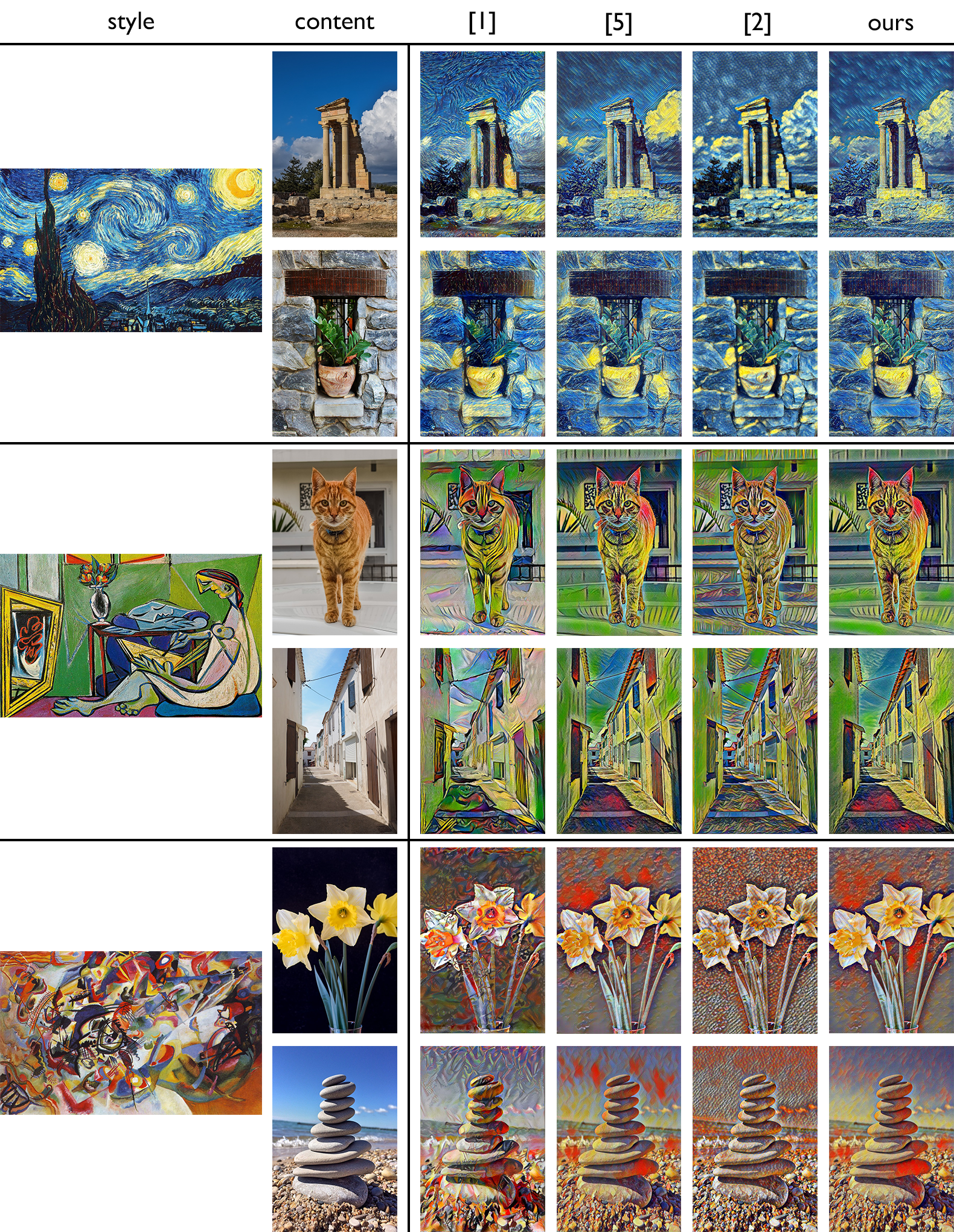}
        \caption{Visual comparison between the methods by Gatys~\etal \cite{gatys2016image}, Johnson~\etal \cite{johnson2016perceptual} (with IN), Liu~\etal \cite{liu2017depth} and ours. The results of the methods by Gatys~\etal \cite{gatys2016image} and Johnson~\etal \cite{johnson2016perceptual} were reproduced based on the original implementations of the authors whereas the results of Liu~\etal \cite{liu2017depth} are retrieved directly from their paper with the authors' permission.}
        \label{fig:results_all_methods}
\end{figure}

Figure~\ref{fig:results_all_methods} presents a visual side-by-side comparison between the results of our method and the results of the methods by Gatys~\etal~\cite{gatys2016image}, Johnson~\etal~\cite{johnson2016perceptual} (with instance normalization (IN)) and Liu~\etal \cite{liu2017depth}. The images generated using the system by Gatys~\etal~\cite{gatys2016image} depict the style patterns of the style images quite well, nevertheless, they fail to preserve the overall structure and 3D layout of the content image. The algorithm of Johnson~\etal~\cite{johnson2016perceptual} enhanced with Instance Normalization does better in terms of preserving the contents of the image and not applying the style patterns evenly throughout the whole image. The same applies for the method of Liu~\etal~\cite{liu2017depth}. Although the results are quite similar, our system manages to better distinguish the objects that are located further away from the objects that are located near the camera. This is more visible in the last two images where our algorithm is capable of identifying the objects at the centre of the image and stylizing them appropriately, avoiding the distribution of uneven brush strokes in the background. 

\begin{figure}[!htb]
	\centering
        \includegraphics[width=1\linewidth]{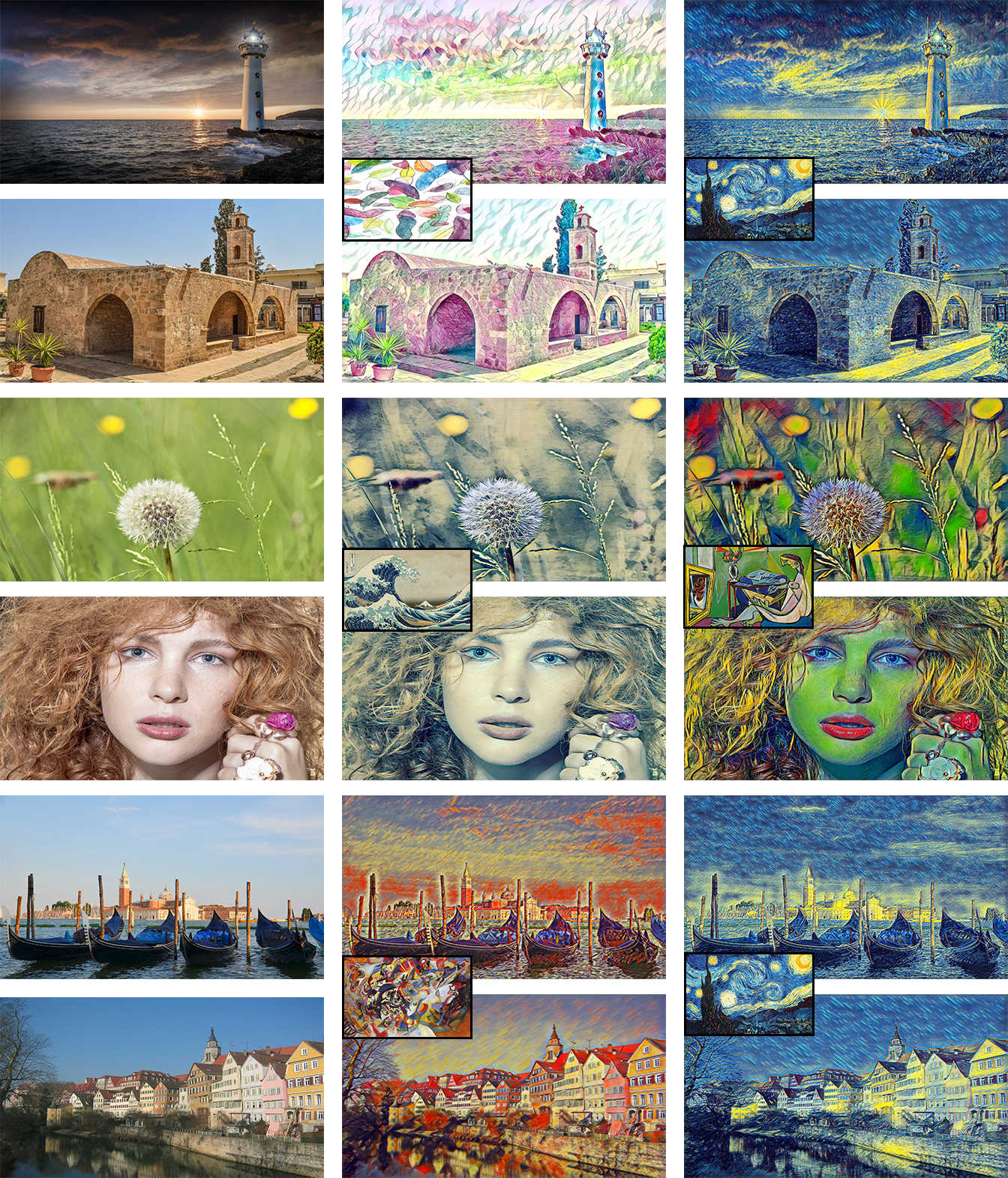}
        \caption{Illustration of our results for different content images and different painting styles. Our approach captures the colour and texture patterns of the style image, while retaining depth information, allowing the objects located at the centre of the image to stand out.}
        \label{fig:our_results}
\end{figure}

More results from models trained on various style images using our approach are demonstrated in Figure~\ref{fig:our_results}. While capturing the colour and texture patterns of the style image, our method also manages to retain the depth and allow the object located at the centre of the image to stand out. This is prominent in the first row of the figure and the third where the church and dandelion head, respectively, are being emphasized.

\begin{figure}[!htb]
	\centering
        \includegraphics[width=\linewidth]{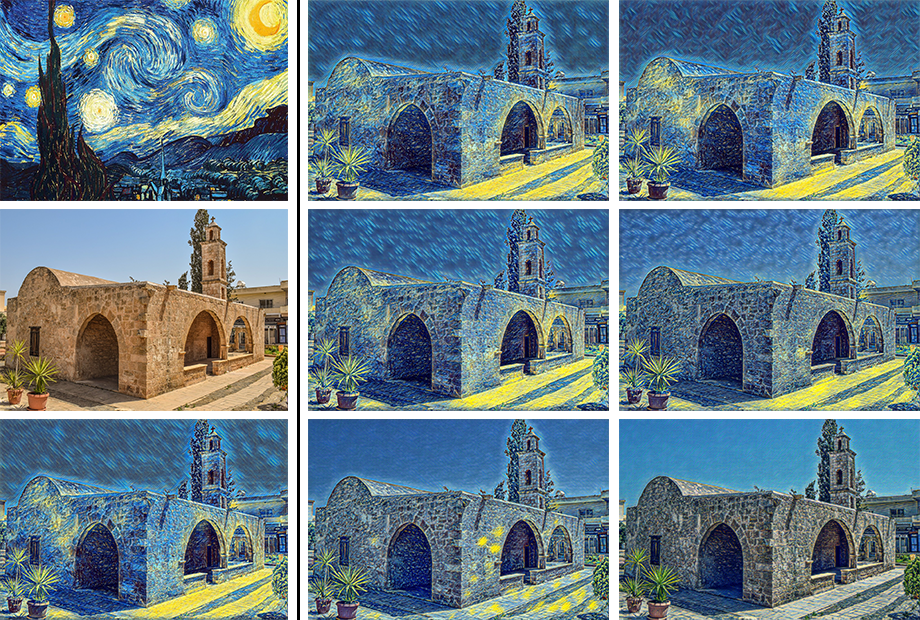}
        \caption{Increasing the depth weight results in more structure being preserved but it captures less of the style patterns. The style, content and the stylized result with no depth information are displayed on the left. Our results are shown in increasing depth weight from top left to the bottom. The top left image is generated with the lowest weight and the bottom right with the highest weight for the depth loss.}
        \label{fig:increasing_depth}
\end{figure}

As Liu~\etal \cite{liu2017depth} have noticed, there is a trade-off between perceptual loss and depth loss. If style loss is emphasized, the style patterns can be captured more accurately, whereas increasing the depth loss weight will preserve more of the contents and spatial layout of the image (Figure~\ref{fig:increasing_depth}).


\subsection{User study}

 

To quantitatively gauge the aesthetic effect of our approach, we conducted a user study. We selected 5 different style images, including the common artistic paintings that are presented amongst previous NST studies' results, and 6 different content images that vary in form, colour, and content, ranging from landscapes to nature photographs and face portraits. The 6 content images are displayed in the left column of Figure~\ref{fig:our_results}, with the 5 style images and resulting styled content images shown in the middle and right columns (with one of the style images repeated). The participants were shown a series of 30 sets of images. One of the images was generated by our algorithm whereas the other three were generated using previous algorithms (Gatys~\etal \cite{gatys2016image}, Johnson~\etal \cite{johnson2016perceptual} and Liu~\etal \cite{liu2017depth}). The order that the images were shown was randomized. For the first half of the questions (1-15), the content and style images were not shown at the start of the question. The same set of 15 questions (16-30) was also presented (in random order) but with the content and style images that were used to generate the results revealed. We reasoned that excluding the content and style images of the questions would allow us to better evaluate the aesthetic effect of the final results, regardless of the generation process. The participants were asked to select the one image of the four stylized images that they visually preferred (i.e. their favourite stylization). We collected results from 20 participants. Examples of the questions shown to the participants can be found on the project's website: \url{https://ioannoue.github.io/depth-aware-nst-using-in.html\#userStudy}.

\setlength{\tabcolsep}{6pt}
\begin{table}
\begin{center}
    
        \begin{tabular}{c|cccc} 
        \hline
        \multicolumn{5}{c}{Most preferable method ratio} \\
        \hline
        Content \& Style & \cite{gatys2016image}  & \cite{johnson2016perceptual} & \cite{liu2017depth}  & ours  \\
        \hline
        
        \textbf{Omitted} & 13.33\% & 13.33\% & 6.67\% & \textbf{80\%}  \\
        \hline
        
        \textbf{Revealed} & \textbf{60\%} & 0\% & 33.33\% & 13.33\%   \\
        \hline 
        \hline
        \multicolumn{5}{c}{Total votes} \\
        \hline
        Content \& Style & \cite{gatys2016image} & \cite{johnson2016perceptual} & \cite{liu2017depth}  & ours  \\
        \hline
        
        \textbf{Omitted} & 22\% & 16.33\% & 18.67\% & \textbf{43\%}  \\
        \hline
        
        \textbf{Revealed} & \textbf{40.33\%} & 15.33\% & 22.33\% & 22\%   \\
        \hline 
        
        \end{tabular}
        
    \caption{The results of the user study. Stylized images using our method are compared against the methods of Gatys~\etal \cite{gatys2016image}, Johnshon~\etal \cite{johnson2016perceptual} (with IN) and Liu~\etal \cite{liu2017depth}. User preferences for the two different types of questions: omitting the content and style images from the question (omitted) and revealing this information (revealed). The first part of the table (Most preferable method ratio) shows the ratio of times a method is chosen as the most preferable method. There are occasions where more than one method is selected as the favourite of all participants -- two or more methods collected the highest (and same) amount of votes. The second part of the table (Total votes) shows the ratio  of the overall amount of votes each method has collected for all the questions}
    \label{tab:user_study}
\end{center}
\end{table}
\setlength{\tabcolsep}{1.4pt}

\begin{figure}[!htb]
\centering
    \includegraphics[width=\linewidth]{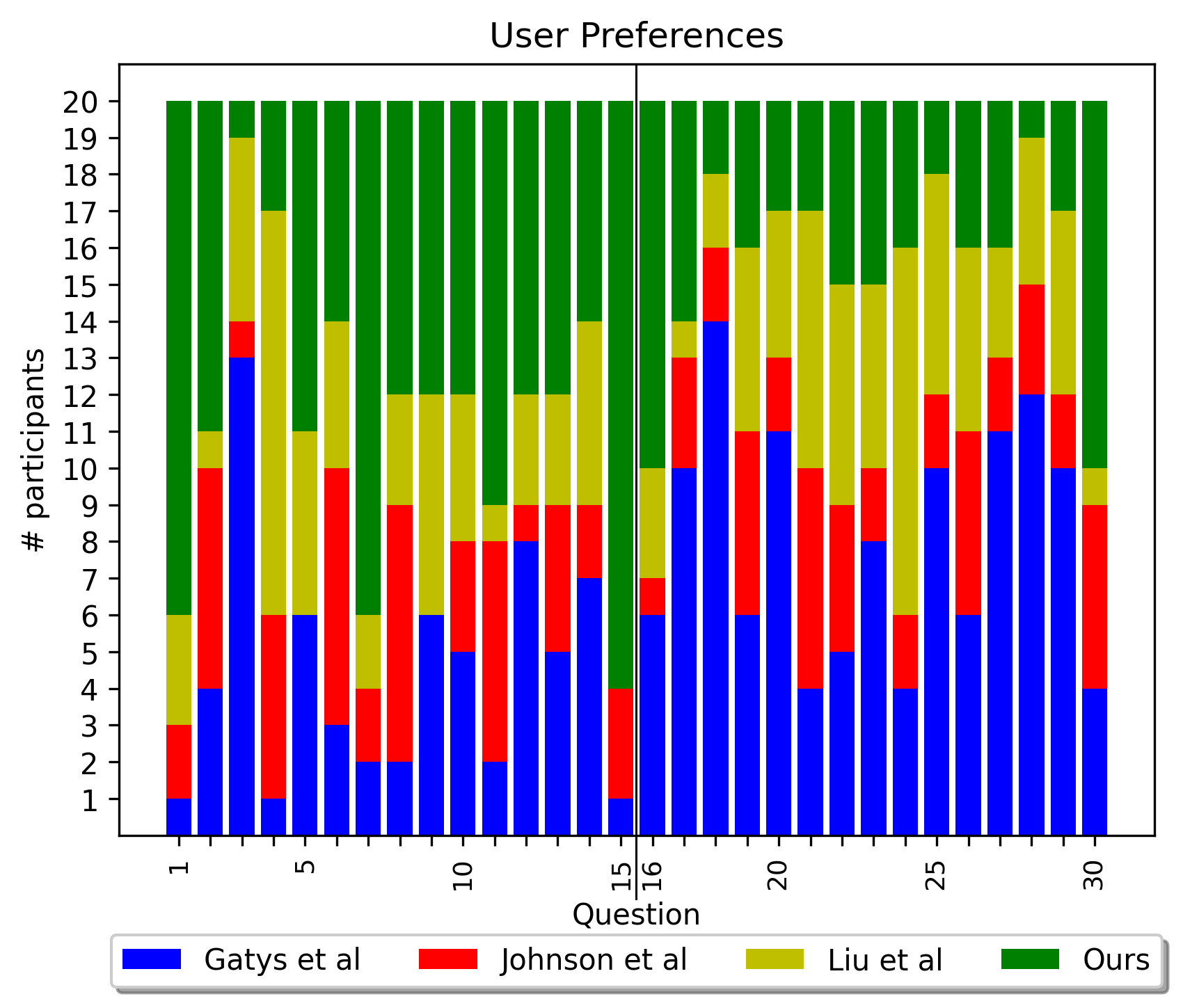}
    \caption{Overall user preferences for all the 30 questions. For questions 1-15 content and style images are omitted whereas for questions 16-30 the content and style images are shown as part of the question.}
    \label{graph:user_preferences}
\end{figure}

The results of the user study are shown in Table~\ref{tab:user_study}. The table shows the ratio of times each method was chosen as the most preferable method or amongst the most preferable methods, and also includes the overall vote distribution for all questions. Our method dominates the user preferences when the content and style images are not revealed to the participants whereas the method of Gatys~\etal \cite{gatys2016image} is the most popular method when the content and style images are included as part of the question. Although the seminal work of Gatys~\etal produces stylizations that capture the style patterns more effectively, our method still performs well and it is superior to the other methods when comparing the final results. This also highlights the significance of the evaluation followed by each NST study, and how each aspect is valued. In this instance, we demonstrate that evaluating only the aesthetics of the results without considering the exact NST operation, the participants prefer different methods compared to when the content and style images are revealed. This is better illustrated by the graph in Figure~\ref{graph:user_preferences}. This shows our method is preferred for the majority of Questions 1-15 where the content and style images are not revealed, perhaps suggesting that when the style image is not shown, depth becomes an important factor in considering an image's quality.


\subsection{Metrics}


Our third evaluation process makes use of metrics inspired by Liu and Zhu's \cite{liu2021structure} evaluation procedure. The aim here is to consider depth and global structure preservation. In terms of structural evaluation of an image, Liu and Zhu suggest Structural Similarity (SSIM) is more compatible with the human visual system (HVS) than peak signal-to-noise ratio (PSNR) and mean-square error (MSE), and thus more suitable for assessing the similarity between a content image and a stylized result. We also use histogram (Hist), average Hash (aHash), and difference hash (dHash) to compare our results with other state-of-the-art methods. The histogram can detect the tonal and colour intensity differences, whereas the image hash \cite{buchner_2021} algorithms analyse the image structure on luminance and can be exploited to identify similar inputs. We use the method of Cai~\etal \cite{cai2018deepdecolor} to perform decolourization on the images before computing the results. This process helps remove any colour (transforming RGB images to grayscale) while preserving the content information. 
This analysis is demonstrated in Table~\ref{tab:quantiative_metrics}. Our method performs better in preserving the structure of the image and the method of Gatys~\etal\cite{gatys2016image} does better in preserving the tonal intensity differences, since their procedure is initialized with a copy of the original content image (which is iteratively being optimized).


\setlength{\tabcolsep}{6pt}
\begin{table}[!htb]
\begin{center}

    \begin{tabular}{ r cccc } 
     \hline\noalign{\smallskip}
    \textbf{} & \textbf{\cite{gatys2016image}}  & \textbf{\cite{johnson2016perceptual}}  & \textbf{\cite{liu2017depth}}  & \textbf{ours}  \\
    \hline\noalign{\smallskip}
    
    \textbf{SSIM} & 0.3951 & {\color{cyan}0.4857} & 0.4176 & {\color{red}0.4905}  \\
    \hline\noalign{\smallskip} 
    
    \textbf{Hist} & {\color{red}0.5518} & {\color{cyan}0.5115} & 0.4555 & 0.4516
   \\
    \hline\noalign{\smallskip} 
    
    \textbf{aHash} & {\color{red}0.9219} & 0.8264 & 0.8316 & {\color{cyan}0.8403}   \\
    \hline\noalign{\smallskip}
    
    \textbf{dHash} & {\color{red}0.8767} & 0.7587 & 0.7378 & {\color{cyan}0.7830}   \\
    \hline 
    
    \end{tabular}
    
    \caption{The average values for SSIM, Hist, aHash and dHash metrics for the methods by Gatys~\etal \cite{gatys2016image}, Johnson~\etal \cite{johnson2016perceptual} (with IN), Liu~\etal \cite{liu2017depth} and our method. The values are measured by comparing the stylized result of each method against the original content image. This is performed for 9 different pairs of images for each method and the averages are calculated. The best results are highlighted with red and the second best with cyan.}
    \label{tab:quantiative_metrics}
\end{center}
\end{table}



Additionally, we perform depth map and saliency map comparison. 
Ideally, the stylized result preserves more of the depth information and structure of the content image. We use the method of Ranftl~\etal \cite{Ranftl2020,Ranftl2021} to compute the depth maps. We measure the structural similarity (SSIM) between the original image's depth map and the stylized result's depth map. Our method manages to preserve more of the depth information.

\setlength{\tabcolsep}{6pt}
\begin{table}
\begin{center}

    \begin{tabular}{rcccc} 
     \hline\noalign{\smallskip}
    \textbf{} & \textbf{\cite{gatys2016image}}  & \textbf{\cite{johnson2016perceptual}}  & \textbf{\cite{liu2017depth}}  & \textbf{ours}  \\
    \hline\noalign{\smallskip}
    
    \textbf{Depth map} & 0.8673 & {\color{cyan}0.8848} & 0.8801 & {\color{red}0.9112}  \\
    \hline\noalign{\smallskip}
    
    \textbf{Saliency map} & 0.4605 & {\color{cyan}0.4723} & 0.4675 & {\color{red}0.5021}   \\
    \hline 
    
    \end{tabular}
    
     \caption{The average SSIM between the depth maps and saliency maps of the original image and the methods by Gatys~\etal \cite{gatys2016image}, Johnson~\etal \cite{johnson2016perceptual} (with IN), Liu~\etal \cite{liu2017depth} and ours. The values are measured by comparing the depth and saliency map of the stylized result of each method against the depth and saliency map of the original content image. This is performed for 10 different pairs of images for each method and the averages are calculated. The best results are highlighted with red and the second best with cyan.}
     \label{tab:depth_sal_map_metrics}
         
\end{center}
\end{table}


Saliency detection is considered an instance of image segmentation and can be used to identify visually predominant regions. The aim of our stylization method is to induce as minor change as possible to the saliency map of the content image, again resulting in more detail preservation. We repeat the structural similarity measurements on the saliency maps to give a more accurate estimate of the results. The method by Jiang~\etal \cite{jiang2013salient} is used to perform saliency detection. The average results for the depth and saliency maps structural similarity are reported in Table~\ref{tab:depth_sal_map_metrics}. Our method performs the best (highlighted with red) in preserving the details of the content as the depth and saliency map of our stylized results are closer to the original image's.

Using an advanced depth estimation method, we show that the results can be significantly improved in comparison with the previous method of Liu~\etal \cite{liu2017depth}. The global structure, depth, and dominant regions are being preserved resulting in aesthetically enhanced results. 

For the depth map comparison, we used the method of Ranftl~\etal \cite{Ranftl2020,Ranftl2021}, whereas in the works of Liu~\etal \cite{liu2017depth} and Cheng~\etal \cite{cheng2019structure} the method of Chen~\etal \cite{chen2016single} was used. Similarly to these methods, the same depth prediction network utilised during training was also used to drive the depth map comparison of the generated results. Future work could consider comparisons using both depth estimation algorithms (\cite{Ranftl2020,chen2016single}).

\subsection{Discussion}

Visual side-by-side comparisons indicate that, similarly to the state-of-the-art methods, our system can properly capture colour and texture patterns of the style image, and in addition, it can produce stylizations that retain depth and allow the main object at the centre of the image stand out. The user study suggests that the effect our algorithm achieves has some positive impact on the aesthetics of the results as it is favoured by the participants. It also raises questions regarding the design of the user study and the form of its presentation, i.e. does showing the style image as part of the question lead a user's thinking about what to consider when interpreting a picture's aesthetic quality? 

Lastly, we have provided a quantitative evaluation based on particular metrics, capable of assessing the depth and structure-preserving capabilities of a method (depicted in Tables~\ref{tab:quantiative_metrics} and \ref{tab:depth_sal_map_metrics}). A small set of images was used for this. A larger evaluation dataset will be considered in future work to increase the validity of the results. Additionally, although SSIM has been chosen as the metric to drive the comparison in depth and saliency maps, different metrics, such as MSE and PSNR, could also be considered.

These three evaluation approaches -- visual side-by-side comparisons, user study and quantitative evaluation -- are all commonly used in previous research, yet the best approach for evaluation is still an open question.


\section{Conclusions}
\label{sec:Conclusions}

We have developed an approach for depth-aware neural style transfer on images. We have demonstrated that our models can effectively stylize 2D images while retaining the depth and global structure of the input content image. Our system replaces the Batch Normalization layers of the generator network with Instance Normalization and employs an advance depth prediction network for the calculation of a depth reconstruction loss. We have shown that Instance Normalization, which is not used in previous depth-aware style transfer methods, improves the quality of the results. In addition, we have shown that a more accurate depth estimation network can help maintain better style contrast across the image and further improve upon the preservation of the hierarchy and depth information.

We have evaluated our results both qualitatively and quantitatively using three different approaches, each of which has been used in previous studies. No single approach prevails. Whilst attempts have been made to propose robust quantitative evaluation procedures that do not rely on user studies \cite{Yeh2018quantitative}, it remains an open question how best to evaluate stylized images. In our future work, we intend to consider the field of computational aesthetics assessment \cite{Zhang2021Comprehensive} and what this might offer to the evaluation process for NST techniques.

\section{Acknowledgements}
This research was supported by the EPSRC [grant number EP/R513313/1].

\printbibliography      

@article{jing2019neural,
  title={Neural style transfer: A review},
  author={Jing, Yongcheng and Yang, Yezhou and Feng, Zunlei and Ye, Jingwen and Yu, Yizhou and Song, Mingli},
  journal={IEEE transactions on visualization and computer graphics},
  year={2019},
  publisher={IEEE}
}

@inproceedings{gatys2016image,
  title={Image style transfer using convolutional neural networks},
  author={Gatys, Leon A and Ecker, Alexander S and Bethge, Matthias},
  booktitle={Proceedings of the IEEE conference on computer vision and pattern recognition},
  pages={2414--2423},
  year={2016}
}

@misc{simonyan2015deep,
      title={Very Deep Convolutional Networks for Large-Scale Image Recognition}, 
      author={Karen Simonyan and Andrew Zisserman},
      year={2015},
      eprint={1409.1556},
      archivePrefix={arXiv},
      primaryClass={cs.CV}
}

@inproceedings{johnson2016perceptual,
  title={Perceptual losses for real-time style transfer and super-resolution},
  author={Johnson, Justin and Alahi, Alexandre and Fei-Fei, Li},
  booktitle={European conference on computer vision},
  pages={694--711},
  year={2016},
  organization={Springer}
}

@inproceedings{ulyanov2016texture,
  title={Texture networks: Feed-forward synthesis of textures and stylized images.},
  author={Ulyanov, Dmitry and Lebedev, Vadim and Vedaldi, Andrea and Lempitsky, Victor S},
  booktitle={ICML},
  volume={1},
  pages={4},
  year={2016}
}

@inproceedings{ulyanov2017improved,
  title={Improved texture networks: Maximizing quality and diversity in feed-forward stylization and texture synthesis},
  author={Ulyanov, Dmitry and Vedaldi, Andrea and Lempitsky, Victor},
  booktitle={Proceedings of the IEEE Conference on Computer Vision and Pattern Recognition},
  pages={6924--6932},
  year={2017}
}

@article{dumoulin2016learned,
  title={A learned representation for artistic style},
  author={Dumoulin, Vincent and Shlens, Jonathon and Kudlur, Manjunath},
  journal={arXiv preprint arXiv:1610.07629},
  year={2016}
}

@inproceedings{chen2017stylebank,
  title={Stylebank: An explicit representation for neural image style transfer},
  author={Chen, Dongdong and Yuan, Lu and Liao, Jing and Yu, Nenghai and Hua, Gang},
  booktitle={Proceedings of the IEEE conference on computer vision and pattern recognition},
  pages={1897--1906},
  year={2017}
}

@inproceedings{huang2017arbitrary,
  title={Arbitrary style transfer in real-time with adaptive instance normalization},
  author={Huang, Xun and Belongie, Serge},
  booktitle={Proceedings of the IEEE International Conference on Computer Vision},
  pages={1501--1510},
  year={2017}
}

@article{chen2016fast,
  title={Fast patch-based style transfer of arbitrary style},
  author={Chen, Tian Qi and Schmidt, Mark},
  journal={arXiv preprint arXiv:1612.04337},
  year={2016}
}

@inproceedings{gu2018arbitrary,
  title={Arbitrary style transfer with deep feature reshuffle},
  author={Gu, Shuyang and Chen, Congliang and Liao, Jing and Yuan, Lu},
  booktitle={Proceedings of the IEEE Conference on Computer Vision and Pattern Recognition},
  pages={8222--8231},
  year={2018}
}

@inproceedings{liu2017depth,
  title={Depth-aware neural style transfer},
  author={Liu, Xiao-Chang and Cheng, Ming-Ming and Lai, Yu-Kun and Rosin, Paul L},
  booktitle={Proceedings of the Symposium on Non-Photorealistic Animation and Rendering},
  pages={1--10},
  year={2017}
}

@article{chen2016single,
  title={Single-image depth perception in the wild},
  author={Chen, Weifeng and Fu, Zhao and Yang, Dawei and Deng, Jia},
  journal={arXiv preprint arXiv:1604.03901},
  year={2016}
}

@article{cheng2019structure,
  title={Structure-preserving neural style transfer},
  author={Cheng, Ming-Ming and Liu, Xiao-Chang and Wang, Jie and Lu, Shao-Ping and Lai, Yu-Kun and Rosin, Paul L},
  journal={IEEE Transactions on Image Processing},
  volume={29},
  pages={909--920},
  year={2019},
  publisher={IEEE}
}

@article{kitov2019depth,
  title={Depth-Aware Arbitrary Style Transfer Using Instance Normalization},
  author={Kitov, Victor and Kozlovtsev, Konstantin and Mishustina, Margarita},
  journal={arXiv preprint arXiv:1906.01123},
  year={2019}
}

@article{ulyanov2016instance,
  title={Instance normalization: The missing ingredient for fast stylization},
  author={Ulyanov, Dmitry and Vedaldi, Andrea and Lempitsky, Victor},
  journal={arXiv preprint arXiv:1607.08022},
  year={2016}
}

@article{Ranftl2020,
	author    = {Ren\'{e} Ranftl and Katrin Lasinger and David Hafner and Konrad Schindler and Vladlen Koltun},
	title     = {Towards Robust Monocular Depth Estimation: Mixing Datasets for Zero-shot Cross-dataset Transfer},
	journal   = {IEEE Transactions on Pattern Analysis and Machine Intelligence (TPAMI)},
	year      = {2020},
}

@article{Ranftl2021,
	author    = {Ren\'{e} Ranftl and Alexey Bochkovskiy and Vladlen Koltun},
	title     = {Vision Transformers for Dense Prediction},
	journal   = {ArXiv preprint},
	year      = {2021},
}

@inproceedings{sanakoyeu2018style,
  title={A style-aware content loss for real-time hd style transfer},
  author={Sanakoyeu, Artsiom and Kotovenko, Dmytro and Lang, Sabine and Ommer, Bjorn},
  booktitle={Proceedings of the European Conference on Computer Vision (ECCV)},
  pages={698--714},
  year={2018}
}

@inproceedings{hu2020aesthetic,
  title={Aesthetic-Aware Image Style Transfer},
  author={Hu, Zhiyuan and Jia, Jia and Liu, Bei and Bu, Yaohua and Fu, Jianlong},
  booktitle={Proceedings of the 28th ACM International Conference on Multimedia},
  pages={3320--3329},
  year={2020}
}

@inproceedings{lin2014microsoft,
  title={Microsoft coco: Common objects in context},
  author={Lin, Tsung-Yi and Maire, Michael and Belongie, Serge and Hays, James and Perona, Pietro and Ramanan, Deva and Doll{\'a}r, Piotr and Zitnick, C Lawrence},
  booktitle={European conference on computer vision},
  pages={740--755},
  year={2014},
  organization={Springer}
}

@inproceedings{silberman2012indoor,
  title={Indoor segmentation and support inference from rgbd images},
  author={Silberman, Nathan and Hoiem, Derek and Kohli, Pushmeet and Fergus, Rob},
  booktitle={European conference on computer vision},
  pages={746--760},
  year={2012},
  organization={Springer}
}

@article{saxena2008make3d,
  title={Make3d: Learning 3d scene structure from a single still image},
  author={Saxena, Ashutosh and Sun, Min and Ng, Andrew Y},
  journal={IEEE transactions on pattern analysis and machine intelligence},
  volume={31},
  number={5},
  pages={824--840},
  year={2008},
  publisher={IEEE}
}

@inproceedings{Butler2012sintel,
title = {A naturalistic open source movie for optical flow evaluation},
author = {Butler, D. J. and Wulff, J. and Stanley, G. B. and Black, M. J.},
booktitle = {European Conf. on Computer Vision (ECCV)},
editor = {{A. Fitzgibbon et al. (Eds.)}},
publisher = {Springer-Verlag},
series = {Part IV, LNCS 7577},
month = oct,
pages = {611--625},
year = {2012}
}

@article{kingma2014adam,
  title={Adam: A method for stochastic optimization},
  author={Kingma, Diederik P and Ba, Jimmy},
  journal={arXiv preprint arXiv:1412.6980},
  year={2014}
}

@article{liu2021structure,
  title={Structure-Guided Arbitrary Style Transfer for Artistic Image and Video},
  author={Liu, Shiguang and Zhu, Ting},
  journal={IEEE Transactions on Multimedia},
  year={2021},
  publisher={IEEE}
}

@INPROCEEDINGS{jiang2013salient,
  author={Jiang, Huaizu and Wang, Jingdong and Yuan, Zejian and Wu, Yang and Zheng, Nanning and Li, Shipeng},
  booktitle={2013 IEEE Conference on Computer Vision and Pattern Recognition}, 
  title={Salient Object Detection: A Discriminative Regional Feature Integration Approach}, 
  year={2013},
  volume={},
  number={},
  pages={2083-2090},
  doi={10.1109/CVPR.2013.271}
 }

@article{cai2018deepdecolor,
	author = {Bolun Cai, Xiangmin Xu and Xiaofen Xing},
	title={Perception Preserving Decolorization},
	journal={IEEE International Conference on Image Processing},
	year={2018}
}

@misc{deliot_guinier_vanhoey_2020,
    title={Real-time style transfer in Unity using deep neural networks},
    url={https://blogs.unity3d.com/2020/11/25/real-time-style-transfer-in-unity-using-deep-neural-networks/},
    journal={Unity Technologies Blog},
    author={Deliot, Thomas and Guinier, Florent and Vanhoey, Kenneth},
    year={2020}
}

@article{ruder2018artistic,
  author    = {Manuel Ruder and
               Alexey Dosovitskiy and
               Thomas Brox},
  title     = {Artistic style transfer for videos and spherical images},
  journal   = {CoRR},
  year      = {2017},
  url       = {http://arxiv.org/abs/1708.04538},
  eprinttype = {arXiv},
  eprint    = {1708.04538},
  bibsource = {dblp computer science bibliography, https://dblp.org}
}

@inproceedings{huang2017real,
  title={Real-time neural style transfer for videos},
  author={Huang, Haozhi and Wang, Hao and Luo, Wenhan and Ma, Lin and Jiang, Wenhao and Zhu, Xiaolong and Li, Zhifeng and Liu, Wei},
  booktitle={Proceedings of the IEEE Conference on Computer Vision and Pattern Recognition},
  pages={783--791},
  year={2017}
}

@misc{gao2018reconet,
      title={ReCoNet: Real-time Coherent Video Style Transfer Network}, 
      author={Chang Gao and Derun Gu and Fangjun Zhang and Yizhou Yu},
      year={2018},
      eprint={1807.01197},
      archivePrefix={arXiv},
      primaryClass={cs.CV}
}

@article{Yeh2018quantitative,
  title={Quantitative evaluation of style transfer},
  author={Yeh, Mao-Chuang and Tang, Shuai and Bhattad, Anand and Forsyth, David A},
  journal={arXiv preprint arXiv:1804.00118},
  year={2018}
}

@article{Zhang2021Comprehensive,
  title={A Comprehensive Survey on Computational Aesthetic Evaluation of Visual Art Images: Metrics and Challenges},
  author={Zhang, Jiajing and Miao, Yongwei and Yu, Jinhui},
  journal={IEEE Access},
  year={2021},
  publisher={IEEE}
}

@misc{buchner_2021,
title={ImageHash},
url={https://pypi.org/project/ImageHash/},
journal={PyPI},
author={Buchner, Johannes},
year={2021}
}

\end{document}